\documentclass[10pt, a4paper]{article}

\usepackage{lrec-coling2024} % this is the new style

\title{Linking Named Entities in Diderot's \textit{Encyclopédie} to Wikidata\\ \vspace*{.5\baselineskip}}

\name{Pierre Nugues} 
\address{Lund University \\
         Lund, Sweden \\
         pierre.nugues@cs.lth.se\\
         \textit{Paper originally published in the Proceedings of the LREC-COLING, 2024}}

\abstract{
Diderot's \textit{Encyclopédie} is a reference work from XVIIIth century in Europe that aimed at collecting the knowledge of its era. \textit{Wikipedia} has the same ambition with a much greater scope. However, the lack of digital connection between the two encyclopedias may hinder their comparison and the study of how knowledge has evolved. A key element of \textit{Wikipedia} is Wikidata that backs the articles with a graph of structured data. In this paper, we describe the annotation of more than 10,300 of the \textit{Encyclopédie} entries with Wikidata identifiers enabling us to connect these entries to the graph. We considered geographic and human entities. The \textit{Encyclopédie}  does not contain biographic entries as they mostly appear as subentries of locations. We extracted all the geographic entries and we completely annotated all the entries containing a description of human entities. This represents more than 2,600 links referring to locations or human entities. In addition, we annotated more than 9,500 entries having a geographic content only. We describe the annotation process as well as application examples. This resource is available at \url{https://github.com/pnugues/encyclopedie_1751}. \\ \newline }

\begin{document}

\maketitleabstract

\section{Introduction}
Diderot's \textit{Encyclopédie} is indisputably a milestone in the intellectual history of  Europe. Published between 1751 and 1772, consisting of 17 volumes of text and 11 volumes of plates, it had a formidable success and, at the same time, spurred controversies and suffered censorship. By its scope and the values it put forward, it defined new standards in the collection and presentation of concepts, sciences, and techniques. The ambition of the \textit{Encyclopédie} was to summarize the knowledge of its time as the authors expressed  it in the \textit{explicit liber} of the last volume:
\begin{quote}
ce Dictionnaire, destiné particulierement à être le dépôt des connoissances humaines. \\
``this Dictionary, intended particularly to be the repository of human knowledge.'' 
\end{quote}

At present, the \textit{Encyclopédie}'s role has been taken over by \textit{Wikipedia} with an analogue endeavor asserted in its Prime Objective \cite{wiki:Wikipedia:Prime_objective}:
\begin{quote}
Imagine a world in which every single person on the planet is given free access to the sum of all human knowledge. That's what we're doing
\end{quote}
No competing multilingual encyclopedia or similar project has \textit{Wikipedia}'s reach, scope, and volume. In addition to be a popular reference for students \citep{thomas2021using,amina2022use}, journalists \citep{messner2011legitimizing}, and academics \citep{dooley2010wikipedia}, it is  a reasonably authoritative and reliable source \citep{giles2005special}. As a consequence, \textit{Wikipedia} is increasingly cited in scientific papers \citep{tomaszewski2016study}.

Beyond the human readers of its websites, \textit{Wikipedia} has a far-reaching influence in information technologies as many NLP applications indirectly encapsulate it. Many large language models and conversational agents (chatbots) use it as their reference knowledge source \citep{longpre2023pretrainer}. BERT \citep{Devlin2019}, a transformer-based encoder, is an example of them. It uses the English version of \textit{Wikipedia} and a collection of books as training corpus. BERT itself serves as the pre-trained model of scores of derived applications: classification, sequence tagging, entailment, question answering, etc.

In this work, we connected a reference work from the past to a reference of today, namely the \textit{Encyclopédie} to \textit{Wikipedia} through named entities. We restricted the links to entries describing geographic and human entities as they are easier to disambiguate and can form a starting point for other categories. 

The contributions of this work are:
\begin{enumerate}
\item We analyzed qualitatively a digital version of the \textit{Encyclopédie} to identify entries that could contain biographies and we determined that biographies were mostly subentries of geographical entries;
\item We extracted the geographical entries using the \textit{Encyclopédie} categories (15,274 entries). We identified all of them containing biographies (841 entries) and we linked all the biographies to a Wikidata identifier (1,715 links);
\item  We annotated additional geographical entries with no biography and we reached a total of more than  10,300 entries. 
\item As an application example of this dataset, we extracted the activities of the human entities, their date of birth and of death from Wikidata as well as the geographical coordinates of the headwords.
%6,719 entries au total
\item Finally, we published the annotated dataset as a JSON file in GitHub\footnote{\url{https://github.com/pnugues/encyclopedie_1751}}.
\end{enumerate}

Using metaphors, Bernard of Chartres and Newton noted that the discovery or creation of new knowledge, most of the time, builds on already existing knowledge. We hope our annotated corpus will enable other researchers to develop tools to improve the understanding of the knowledge transmission process.

\section{Previous Work}
\paragraph{Entry Linking and Comparison with Wikipedia.}
There are few works linking old encyclopedias or dictionaries to Wikidata identifiers. \citet{nugues2022connecting} is an example, where the author linked manually 20,245 entries to Wikidata. Out of these links, about 7,600 apply to human entities.

More generally, \citet{10.1093/llc/fqac083} gives a list of applications using Wikidata in digital humanities. A few of them use it for entity linking, but none specifically for encyclopedias.

Studies contrasting \textit{Wikipedia} with encyclopedias written by experts often focus on differences with \textit{Encyclopædia Britannica}. They usually identify equivalent entries from both sources using automatic string matching or a manual examination and compare the respective texts.

\citet{greenstein2014experts} tried to assess the respective biases of these two encyclopedias for ``contested knowledge'' in US politics. They extracted  3,918 pairs of matching articles and compared their content. Using a set of phrases that are typical of Democrats or Republicans, the authors found that \textit{Wikipedia} was  leaning more towards Democratic views. In the matching procedure, the authors kept titles that were identical or nearly identical and checked them manually. 

\citet{samoilenko2018don} carried out an analysis of the history of nations on both encyclopedias and found that \textit{Encyclopædia Britannica} leant more on cultural phenomena and geography, while \textit{Wikipedia} focused more on wars and popular events. The authors relied on date extraction, string matching, and lexical statistics. 

While lexical statistics using string matching is a respectable baseline technique, it can lead to inaccuracies. Names of countries, people, or concepts are often ambiguous. A country or a location can have multiple names and a name can designate multiple things. In comparison, linking the entries or the words to unique identifiers, such as those in Wikidata, disambiguates the words and connects them univocally to databases \citep{6823700}.

\paragraph{Linking Entries of the \textit{Encyclopédie}.}
Systematic linking of entries in the \textit{Encyclopédie} are less frequent and, to the best of our knowledge, do not relate them to Wikidata or \textit{Wikipedia}.  \citet{Moncla2019} identified location strings in the entries and built a graph of them. They did not link the strings to unique identifiers such as those in Wikidata.

\citet{10.3917/lf.214.0059} sampled 108 geographical entries of the  \textit{Encyclopédie} and compared them manually to equivalent entries from another French encyclopedia of the same time: The Trévoux dictionary. They identified the sources the authors used to write their entries and outlined the disparate structure of the entries in the \textit{Encyclopédie}. They did not link the headwords to unique identifiers.

\citet{10.1145/3149858.3149861} applied an automatic analysis to all the geographical entries of the \textit{Encyclopédie} with a gazetteer. They validated manually the annotation on 100 entries and found that  their geographical parser could annotate only 40\% of them. This hints at a necessary manual validation of all the links.

\paragraph{Digitization.}
It would be extremely tedious, if not impossible, to systematically link the entries of an encyclopedia without a digital version of it. For the \textit{Encyclopédie}, this digitization was not a trivial task given its size and the lack of dedicated optical character recognition (OCR) tools to transcribe XVIIIth century French from image scans. 

The digital version of the \textit{Encyclopédie} is the result of three main projects. \newcite{morrissey1998encyclopedie} produced a first OCRed version at the University of Chicago that they complemented with a substantial manual correction. Wikisource, Wikimedia's free library, made this text available so that anonymous volunteers could edit the text and remove scores of transcription errors. Finally, the ENCCRE project\footnote{\url{http://enccre.academie-sciences.fr/encyclopedie/}} carried out a new digitization of the original edition of the \textit{Encyclopédie} and, starting from the Wikisource text, added an XML-TEI markup and interlinks to the 74,125 entries \cite{guilbaud2017enccre,guilbaud2013entrer}. We used this ENCCRE version in our work.

\section{Collection of the Corpus}
We collected all the entries with a geographic category from the ENCCRE website. In total, we obtained 15,274 documents representing a fifth of the entries (20.6\%). The overwhelming majority of these entries corresponds to locations, while a few of them describe geographic concepts such as \textit{fleuve} `river' or \textit{montagnes} `mountains'.

\begin{figure}[tbh]
\includegraphics[width=1.0\columnwidth]{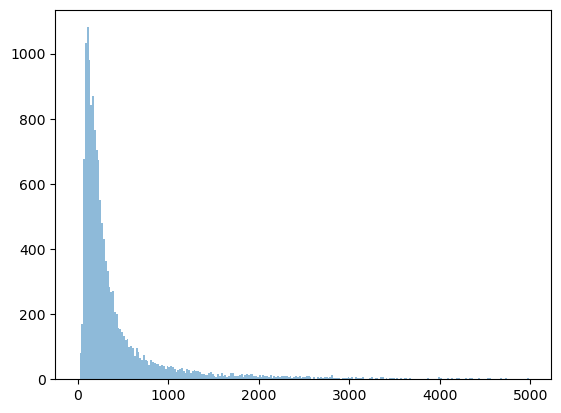}
\centering
\caption{Frequency histogram of the distribution of entries by length in number of characters}
\label{fig:histo_letters}
\end{figure}

Figure \ref{fig:histo_letters} shows the length distribution of the entries by length in number of characters. The median is of 232 characters including the headwords and the mean is of 700. We observe that the distribution is heavily skewed to the left, meaning that most entries consist of rather short texts. The shortest ones being cross references and small towns with no more than 40 characters. The three longest entries are \textit{Fontaine} `Spring', \textit{Paris}, and \textit{Géographie physique} `Physical geography' with respectively 146,589, 115,339, and 91,641 characters.   

We tokenized the entries of the geographic category to extract the words and numbers. We used the Unicode regular expression: \verb='\p{L}+|\p{N}+'= and we obtained 1,886,628 words in total with a median of  42 words per entry and a mean of 123.5.

\section{Entry Structure and Annotation}
\paragraph{Manual Annotation.}
We started with an automatic search of entities using the headwords and the text of the entries. We created a program to query \textit{Wikipedia}, wikidata, and the web with the duckduckgo search engine. We extracted the wikidata identifiers from the pages it returned. This provided us with short lists of candidates for each entry. Unfortunately, this procedure returned many wrong links and missed of lot of entities. We replaced it with a systematic manual annotation that we verified with other encyclopedias or dictionaries from the XVIIIth or XIXth centuries. 

The annotation rapidly revealed quite challenging for many cases. The geographic descriptions are often short, as shown in Fig. \ref{fig:histo_letters}, sometimes imprecise, and for a few rare cases, inexact. The names or spelling of the cities, regions, or countries may have changed, making the identification difficult. Sometimes, some concepts from the XVIIIth century or locations simply do not exist in Wikidata. 

The philosophers of the Enlightenment were still steeped in the Classics. This explains that many entries in the \textit{Encyclopédie} refer to locations mentioned by ancient geographers. These locations represent often tricky annotation cases as many of them have never been properly identified. The \textit{Physcus} headword is an example of them, where the entry text mentions seven locations from Ancient Greece geographers:
\begin{quote}
PHYSCUS, (Géog. anc.) il y a plusieurs lieux de ce nom ; savoir, 1°. Une ville de l’Asie mineure, [...] 2°. Une ville des Ozoles de la Locride, Plutarque en parle dans ses questions grecques ; 3°. une ville de la Carie, selon Etienne le géographe ; 4°. une ville de la Macédoine, selon le même auteur ; 5°. il donne aussi ce nom à un port de l’île de Rhodes ; 6°. un fleuve aux environs de l’Assyrie, suivant un passage de Xénophon, \textit{l. II. de Cyri exped.} cité par Ortelius ; 7° une montagne d’Italie dans la grande Grece, près de Crotone, selon Théocrite. \textit{Idyl. 4.} 

``PHYSCUS, (Anc. Geog.) there are several places of this name; namely, 1°. A city in Asia Minor, [...] 2°. A city of the Ozoles of Locris, Plutarch speaks of it in his Greek questions; 3°. a city in Caria, according to Stephanus of Byzantium; 4°. a city in Macedonia, according to the same author; 5°. he also gives this name to a port on the island of Rhodes; 6°. a river in the vicinity of Assyria, following a passage from Xenophon, \textit{l. II. by Cyri exped.} cited by Ortelius; 7° a mountain in Italy in Greater Greece, near Crotone, according to Theocritus. \textit{Idyl. 4.}''
\end{quote} 
To the best of our effort, we could not identify items 2, 3, 5 and 7. In such cases, we used the \verb=Q0= identifier which does not exist in Wikidata to mark the unresolved entities. We annotated \textit{Physcus} with this list of Wikidata identifiers:
\texttt{["Q209908", "Q0", "Q0", "Q60792888", "Q0", 
 "Q7826058", "Q0"]}. As of today, we annotated 217 entries of the dataset with at least one \verb=Q0= identifier.

\paragraph{Human Being Subentries.}
 In the \textit{Encyclopédie}, no entry headword corresponds to a human being. This does not mean that it does not contain biographies. The structure is somehow different to that of contemporary encyclopedias as they mostly show inside geographical entries as separate subentries. The following excerpt of the entry for the city of Grenoble gives an example of such biographies:
\begin{quote}
GRENOBLE, \textit{Gratianopolis}, (\textit{Géogr.}) ancienne ville de France, capitale du Dauphiné, [...].

On met au nombre des jurisconsultes dont Grenoble est la patrie, Pape (Guy), qui mourut en 1487 ; son \textit{recueil de décisions des plus belles questions de droit}, n’est pas encore tombé dans l’oubli.

M. de Bouchenu de Valbonnais, (Jean Pierre Moret) premier président du parlement de \textit{Grenoble}, né dans cette ville le 23 Juin 1651, mérite le titre du \textit{plus savant historiographe de son pays}, par la belle \textit{histoire du Dauphiné}, qu’il a publiée en trois \textit{vol. in fol.} il est mort en 1730, âgé de 79 ans. [...]

``GRENOBLE, \textit{Gratianopolis}, (\textit{Geogr.}) ancient city of France, capital of Dauphiné [...]

Among the jurisconsults whose homeland is Grenoble are Pape (Guy), who died in 1487; his \textit{collection of decisions on the finest questions of law}, has not yet been forgotten.

Mr. de Bouchenu de Valbonnais, (Jean Pierre Moret) first president of the parliament of \textit{Grenoble}, born in this city on June 23, 1651, deserves the title of \textit{the most learned historiographer of his country}, by the beautiful \textit{history of the Dauphiné}, which he published in three \textit{vols. in fol.} he died in 1730, aged 79. [...]''
\end{quote}

We annotated the \textit{Grenoble} entry with three Wikidata identifiers:
\begin{enumerate}
\item \verb=Q1289=, the city of Grenoble;
\item \verb=Q41617345=, Gui Pape (c. 1402-1487), French jurist-consult; and
\item \verb=Q3169582=, Jean-Pierre Moret de Bourchenu (1651-1730), French historian.
\end{enumerate}
The entry headword, \textit{Grenoble} in our example, is often the place of birth of the human entities or where s/he died. This enabled us to crosscheck the locations and the persons, and ensure the correctness of the annotation.

A few other biographies occur under doctrine or philosophy headwords like \textit{Platonism} or \textit{Aristotelianism} that include the lives of their creators, here respectively Plato and Aristotle. They are less frequent than biographies associated with location headwords though. As we could not identify them as easily as with the entries on geography, we set them aside.

In this work, we decided to first focus on the locations with biographies. We carried out an exhaustive manual annotation of all the human entities of the \textit{Encyclopédie} when they occurred in entries on geography. We also annotated the corresponding headwords as well as the other location subentries it could contain. We additionally annotated more than 9,500 geographic entries with no biography.

\section{Format of the Annotated Dataset}
We stored the annotated dataset as a list of dictionaries, where each dictionary represents an entry. We used the JSON format, where a dictionary contains: The entry headword, for instance \textit{Grenoble;} its ENCCRE identifier, \verb="v7-1475-0"= for \textit{Grenoble;} the complete text of the entry; and the list of Wikidata identifiers with the \verb='qid'= key, starting with the headword and followed by the subentry headwords. These subentries can be other locations or biographies, for instance for \textit{Grenoble:} \verb=["Q1289", "Q41617345", "Q3169582"]=.

\section{Analyzing the Entities}
\begin{figure}[tbh]
\centering
\includegraphics[width=\columnwidth]{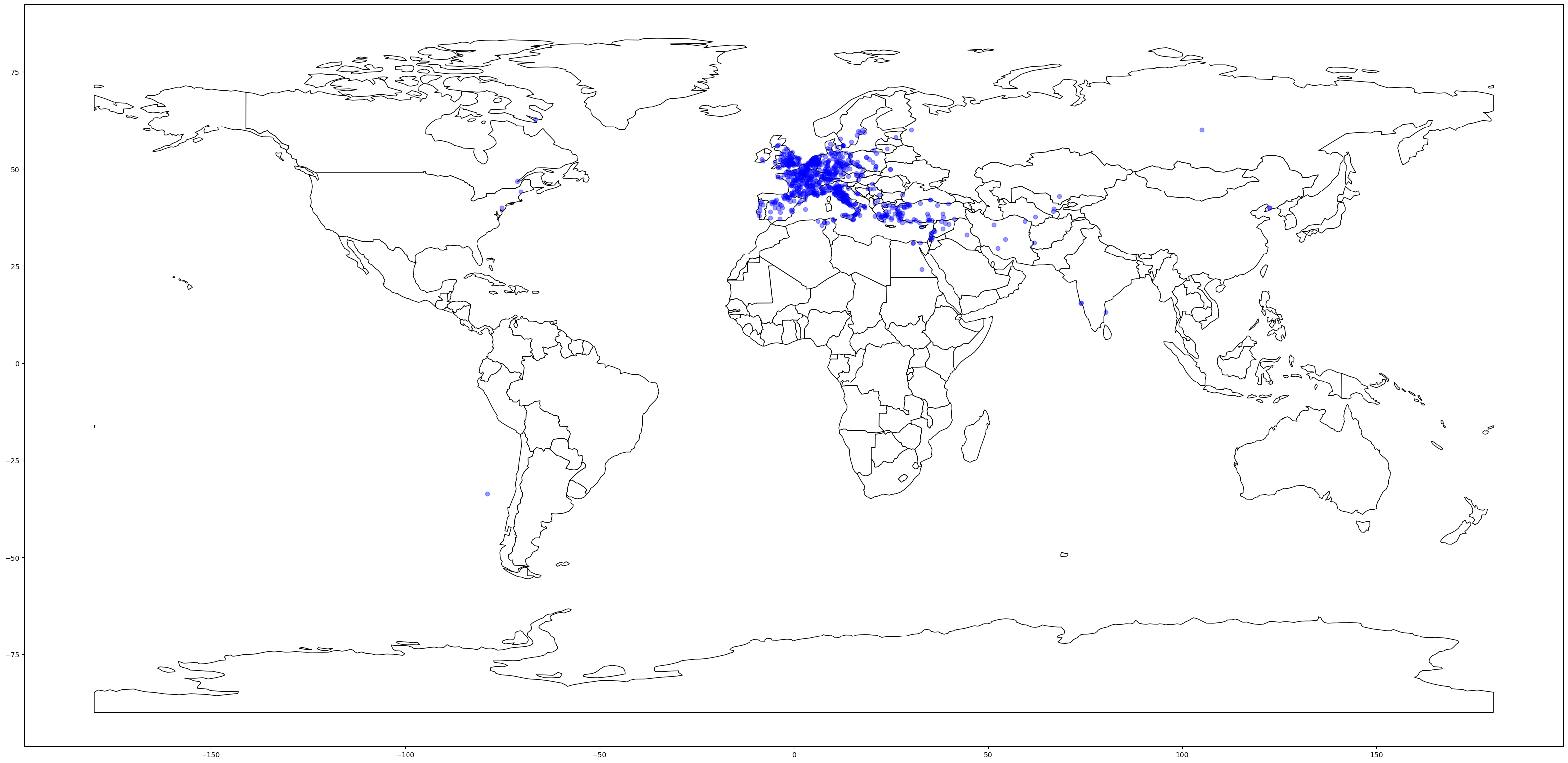}
\caption{Locations of the \textit{Encyclopédie} headwords where a human being is mentioned}
\label{fig:lieuxsources}
\end{figure}

Linking the entities to Wikidata enables us to apply a fine-grained analysis of the geographical horizons of the \textit{Encyclopédie} authors. Using SPARQL queries, we extracted the locations of the headwords when the entry mentioned human entities, frequently their places of birth or death. Figure \ref{fig:lieuxsources} shows the results, where, with only a few exceptions, the human entities are cited in locations in Europe or in the Near East.

\begin{figure}[tbh]
\centering
\includegraphics[width=1.0\columnwidth]{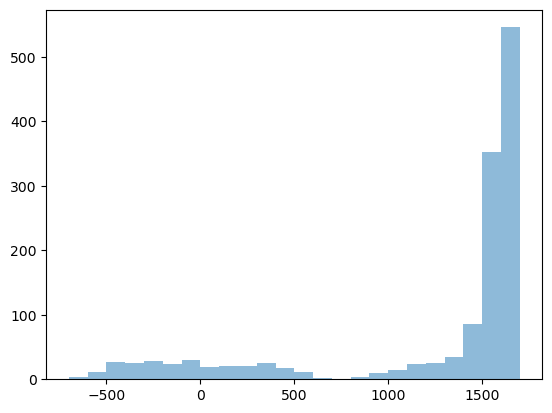}
\caption{Dates of deaths of the people mentioned in the \textit{Encyclopédie} between -700 and 1700}
\label{fig:ddeces}
\end{figure}

Using SPARQL again, we extracted the dates of birth and death of the human entities. Figure~\ref{fig:ddeces} shows those of the death dates, where we observe that the \textit{Encyclopédie} writers focused on Classics and even more on people who died within the two centuries preceding its publication. We also note a sort of black hole between 500 and 1400. 

For most human entities, Wikidata provides a list of one or more activities, called occupations.  For a few lesser-known entities, there is no documented activity though. These activities are significant as they shed a light on the selection process of the entries and who deserves to be in this encyclopedia.  A possible guess would be the kings, queens, emperors, and other rulers in the then era of  ``absolute monarchy'' in France. Surprisingly, this category comes only at the 7th rank under the name \textit{politician} (statesmen/women) in Wikidata; see Table \ref{fig:occupations}. The top five consists of writers, theologians, poets, philosophers, and historians. We may posit that this is a sign of the \textit{Age of Enlightenment} or the  \textit{Republic of Letters}. This also possibly reflects a change of perspective in the history narratives.

\begin{table}[tbh]
\centering
\begin{tabular}{llc}
\hline
\textbf{Qid}& \textbf{Description}&\textbf{Count}\\
\hline
Q36180& Writer &545\\
 Q1234713 &Theologian& 285\\
 Q49757&Poet& 281\\
 Q4964182&Philosopher&249\\
 Q201788&Historian&245\\
 Q1622272&University teacher&224\\
 Q82955&Politician&199\\
 Q250867&Catholic priest&144\\
 Q333634&Translator&125\\
 Q170790 &Mathematician&123\\
 \hline
 \end{tabular}
 \caption{Occupations of the human entities extracted from Wikidata. Note that an entity may have more than one occupation}
 \label{fig:occupations}
\end{table}

\section{Future Work and Conclusion} 
In this work, we described a language resource, where we linked geographic and human entities in the XVIIIth century \textit{Encyclopédie} to Wikidata. In addition to all the geographic headwords with human biographies, we annotated more than 9,500 supplementary geographic locations and concepts. As future work, we plan to annotate the remaining geographic entries, about 4,900. 

To make this subsequent annotation easier, we partitioned the world into 32 regions. We annotated all the remaining entries with a \verb=qid_region= key based on the words in the description of the first subentry such as \textit{Italy} in:
\begin{quote}
ASTRUNO, montagne d’Italie, au royaume de Naples, près de Puzzol ; [...]

``ASTRUNO, mountain of Italy, in the kingdom of Naples, near Puzzol; [...]''

\end{quote}
The three most frequent regions in the JSON dataset are Greece, Italy, and Africa. Note that these regions do not necessarily reflect the current political division of the world. 

As NLP resource, we hope this dataset can help train and assess entity solvers on historic text for both geographic locations and human beings. It could also serve digital humanity research by connecting the \textit{Encyclopédie} to the Wikidata graph. The links should enable researchers to extract more data, for instance to rebut or confirm hypotheses on the \textit{Encyclopédie} content and facilitate further connections with other data sources. 

We believe this work could be adapted to other encyclopedias like the geographic dictionaries that served as main sources for the \textit{Encyclopédie} entries: the \textit{Grand dictionnaire géographique et critique} by Bruzen de La Martinière, 1726-1739, and the \textit{Dictionnaire géographique-portatif} by Vosgien, 1747 \citep{10.3917/lf.214.0059}, other encyclopedias of the same time in French like the \textit{Trévoux} dictionary, or in other languages like the \textit{Universal-Lexicon} from 1731 to 1754 in German.

%The annotated resource will be available for download when the paper is accepted.

\section{Acknowledgments}
We would like to thank the anonymous reviewers for their suggestions and comments.

This work was partially supported by \textit{Vetenskaprådet}, the Swedish Research Council, registration number 2021-04533.

\nocite{*}
\section{Bibliographical References}\label{sec:reference}

\bibliographystyle{lrec-coling2024-natbib}
\bibliography{biblio}

\end{document}